\pdfoutput=1

\documentclass[11pt]{article}

\usepackage[review]{emnlp2022}

\usepackage{times}
\usepackage{latexsym}

\usepackage[T1]{fontenc}

\usepackage[utf8]{inputenc}

\usepackage{microtype}
\usepackage{graphicx}
\usepackage{subfigure}
\usepackage{booktabs} 

\usepackage{amsfonts}
\usepackage{stfloats}
\usepackage{ulem}

\usepackage{amsmath}

\usepackage{adjustbox}
\usepackage{multirow}
\usepackage{array}
\usepackage{float}
\usepackage{caption}
\usepackage{balance}
\usepackage{flushend}

\title{Prompt Tuning for Generative Multimodal Pretrained Models}

\author{Hao Yang$^{*}$, Junyang Lin$^{*}$, An Yang, Peng Wang, Chang Zhou, Hongxia Yang \\ 
DAMO Academy, Alibaba Group \\ 
\{yh351016, junyang.ljy, ya235025, zheluo.wp, ericzhou.zc, yang.yhx\}@alibaba-inc.com}

\begin{document}
\maketitle
\begin{abstract}
Prompt tuning has become a new paradigm for model tuning and it has demonstrated success in natural language pretraining and even vision pretraining. 
In this work, we explore the transfer of prompt tuning to multimodal pretraining, with a focus on generative multimodal pretrained models, instead of contrastive ones. 
Specifically, we implement prompt tuning on the unified sequence-to-sequence pretrained model adaptive to both understanding and generation tasks. 
Experimental results demonstrate that the light-weight prompt tuning can achieve comparable performance with finetuning and surpass other light-weight tuning methods. 
Besides, in comparison with finetuned models, the prompt-tuned models demonstrate improved robustness against adversarial attacks. 
We further figure out that experimental factors, including the prompt length, prompt depth, and reparameteratization, have great impacts on the model performance, and thus we empirically provide a recommendation for the setups of prompt tuning. 
Despite the observed advantages, we still find some limitations in prompt tuning, and we correspondingly point out the directions for future studies. Codes are available at \url{https://github.com/OFA-Sys/OFA}

\end{abstract}

\renewcommand{\thefootnote}{\fnsymbol{footnote}}
\footnotetext[1]{Equal Contribution.}

\section{Introduction}

Recent years have witnessed the great success of large-scale pretraining based on large models and big data in natural language processing (NLP)~\citep{gpt, bert, xlnet, roberta, t5, gpt3} and computer vision~\citep{simclr, simclrv2, moco, mocov3, beit, mae}. 
Mirroring the success of BERT-like models~\citep{bert}, researchers have found that pretraining can level up the downstream performance of cross-modal representation learning algorithms by a large margin~\citep{uniter, vilbert, vl-bert, lxmert}. 
Recent advances show that this idea is compatible with sequence-to-sequence (Seq2Seq) learning, and the Seq2Seq-based multimodal pretrained model can adapt to both understanding and generation tasks, and even achieve the state-of-the-art performance in a series of downstream tasks~\citep{vlt5, simvlm, ofa}.


Despite the great success of large-scale pretrained models across multiple domains, training such models requires a large amount of computation costs. 
The conventional finetuning is though effective in gaining high performance yet suffers from low training efficiency, especially when the pretrained model is of large scale in model size. 
\citet{gpt3} introduced the idea of prompt to encourage the model to generate the correct answer with a manual prompt of task instruction or a demonstration of several task examples, without further training to tune the model parameters. 
This is often regarded as ``in-context learning'', as the model generates responses based on the given context. 
It helps large-scale pretrained language models achieve unprecedented performance in few-shot and zero-shot learning~\citep{gpt3, palm, t0, flan}. 
Inspired by this idea, researchers have moved forward to a new paradigm called \textbf{prompt tuning}~\citep{prefix_tuning, p_tuning, prompt_tuning, prompt_tuning_survey}. 
In comparison with finetuning, prompt tuning only tunes pretrained models by a trivial amount of parameters (e.g., 1\%). 
Prompt tuning freezes most parameters of the pretrained model and only tunes several prompt embeddings, as well as the output layer if necessary. Recent advances have shown that prompt tuning can help pretrained models achieve comparable performance with finetuning across different NLP downstream tasks, including natural language understanding and generation~\citep{p_tuning_v2, unified_view}. 
Such significant achievements have attracted attention of the research community of large pretrained models.

In the domains other than NLP, recent studies have also demonstrated the effectiveness of prompt tuning. 
\citet{vpt} demonstrated that visual prompt tuning could surpass finetuning across a series of tasks, and its advantages in training efficiency were significant. 
In cross-modal representation learning, the research in prompt tuning mainly focuses on the CLIP-like models~\citep{clip}. 
CLIP is a contrastive-learning-based multimodal pretrained model, pretrained on large-scale image-text pairs. 
CLIP is able to achieve outstanding performance in zero-shot image classification by turning labels to textual prompts with manual prompt templates. 
To enhance the performance, \citet{clip} proposed prompt ensembling by handcrafting a number of prompt templates. 
However, as creating hard prompts is tedious, researchers turned to the application of soft prompts for CLIP~\citep{denseclip, coop, cocoop} or the incorporation of adapters~\citep{clip_adapter, tip_adapter}. 
Except for the implementation on CLIP-like models, another line of work is the application of image prompts to pretrained language models for multimodal representation learning~\citep{cpt, frozen}. 
Though the large-scale pretrained laguage model is frozen in the process of downstream transfer, it can adapt to the few-shot learning scenarios of multimodal downstream task. 
Be that as it may, prompt tuning for the popular generative multimodal pretrained models, including BERT-like models and encoder-decoder pretrained models for cross-modal representation learning, is still unexplored. 
\citet{pevl} matched the tuning paradigm to the pretraining one with manual prompts. 
Yet it is still unknown whether the light-weight prompt tuning can also be effective for the generative multimodal pretrained model. 

This work fills in the void and takes the lead to explore prompt tuning for the generative multimodal pretrained models. 
The objective of this study is to investigate whether prompt tuning is effective for the downstream transfer of generative multimodal pretrained models, and how it benefits large pretrained models in comparison with the conventional finetuning. 
To be specific, we implement the simple but effective prefix tuning, one of the most popular prompt tuning methods, on the generative multimodal pretrained model. Prefix tuning owns the advantage of simplicity but at the same time is able to achieve remarkable performance in either natural language understanding or generation~\citep{prefix_tuning, p_tuning_v2}. 
In comparison with finetuning, the number of tunable parameters for prompt tuning is much smaller (\textasciitilde1\%), leading to fewer computation costs, e.g., memory. 

Through extensive experiments we observe that the light-weight prompt tuning is able to help the pretrained model achieve comparable performance with finetuning across $4$ multimodal downstream tasks, spanning from understanding to generation.
To analyze the difference between finetuning and prompt tuning, we follow the assumption that prompt tuning with most parameters in the pretrained model frozen should induce model robustness. 
We experiment on the tuning methods with adversarial attack and observe phenomena consistent with the hypothesis. 
To make a step further, this study delves into the implementation details and investigate whether experimental factors like the prompt length, prompt depth, and reparameterization could saliently influence the final downstream performance. 
We find that in general a longer prompt length (longer than $20$ tokens) is a preferable choice, and our experiments show that $64$ should be favored in most cases as a longer prompt sequence will not only increase the computation costs but also incur performance degradation.
Also, we show that reparameterizaton with additional trainable parameters cannot introduce significant improvements in downstream performance. 
Finally, we reflect on the method and illustrate its defects of computation costs and training instabilities, and correspondingly, we point out some directions for the future work.

In the following sections, we briefly review the related work, deliver an introduction to prompt tuning for generative multimodal pretrained models, and report the experimental results and analysis. Lastly, we discuss the problems of prompt tuning in this scenario, point out the future work, and finally conclude this work. 

\section{Related Work}

In this section, we include the review of multimodal pretraining as well as prompt tuning. We first review the studies in the two main lines of multimodal pretraining, namely generative pretraining and contrastive pretraining, and we then review researches of prompt-based learning in both NLP and cross-modal representation learning. 

\subsection{Multimodal Pretraining}

The rise of vision \& language pretraining started from the transfer of BERT~\citep{bert} to cross-modal representation learning. A series of studies~\citep{vilbert, vl-bert, lxmert, uniter, unicoder-vl} introduced BERT to multimodal pretraining. 
The encoder-decoder framework for multimodal pretraining has recently raised attention, as a number of encoder-decoder models achieved state-of-the-art performance in the cross-modal understanding and generation tasks~\citep{simvlm, ofa, coca}. 
Besides, such framework allows the unification of tasks to sequence-to-sequence learning format and thus allows multitask pretraining with manual prompts~\citep{vlt5, ofa}. 
This leads to our motivation that prompt tuning should be a perfect combination with the recent unified multimodal pretrained model and it can unleash the power of pretrained models with much fewer computation costs than the conventional finetuning. 

Another trend in multimodal pretraining is contrastive learning. 
The most typical constrastive pretrained model is CLIP~\citep{clip}. 
It uses a Vision Transformer (ViT)~\citep{vit} or ResNet~\citep{resnet, tan2019efficientnet} as the image encoder and a transformer model as the text encoder, and trains the two encoders jointly with contrastive loss~\citep{infonce}. 
Note that this model is pretrained on extremely large-scale data of image-text pairs. 
Following CLIP, a series of studies demonstrated the success of this route of contrastive-learning-based pretraining on large-scale data~\citep{align, filip}. 
CLIP can achieve remarkable performance in cross-modal retrieval. What makes it really attractive is its strong performance in zeroshot classification with prompt ensembling, i.e., ensembling the outputs of the model with a handful of handcrafted prompts as the inputs. 
This started the research of prompt in multimodal pretraining. 

\subsection{Prompt-based Learning}

\citet{gpt3} illustrated that large-scale pretrained models can learn from the context and perform few-shot and zero-shot learning with the prompts of  task instruction or a few task examples. 
Instead of using hard prompts by handcrafting, \citet{prefix_tuning} demonstrated that only tuning soft prompt embeddings at each layer is sufficient for the pretrained model to achieve competitive performance in natural language generation, and later a number of studies showed that prompt tuning can be essentially effective for low-resource scenarios~\citep{p_tuning, ppt, black_box} and it can even achieve comparable performance with finetuning~\citep{prompt_tuning, p_tuning_v2}. 
Following this trend, a series of modification to prompts and adapters~\citep{know_prompt, unified_view,ipt,bbtv2} for improvements in performance or training efficiency have emerged and made prompt tuning a heated topic in the whole NLP community. 

Recent prompt tuning methods for multimodal pretrained models mostly serve for CLIP-like models~\citep{coop, cocoop, denseclip}. Similarly, researchers tried to incorporate adapters to CLIP and also achieved satisfactory performance~\citep{clip_adapter, tip_adapter}. 
Except for prompt tuning for CLIP-like models, another line of work explored visual prompts for frozen language models. 
\citet{frozen} showed that when there is a powerful large pretrained language model, a visual encoder for prompt tuning is sufficient for multimodal few-shot learning. 
To take a step forward, \citet{flamingo} proposed Flamingo, a colossal multimodal model that enables in-context learning. 
It could achieve state-of-the-art performance in a series of cross-modal downstream tasks in either few-shot or full-shot learning scenarios. 
Such tremendous success indicates the strong potential of prompt tuning in multimodal pretraining. 
In this work, we focus on an unexplored topic, prompt tuning for generative multimodal pretrained model. 


\section{Method}

\begin{figure*}[t]
    \centering
    \includegraphics[width=1.0\linewidth]{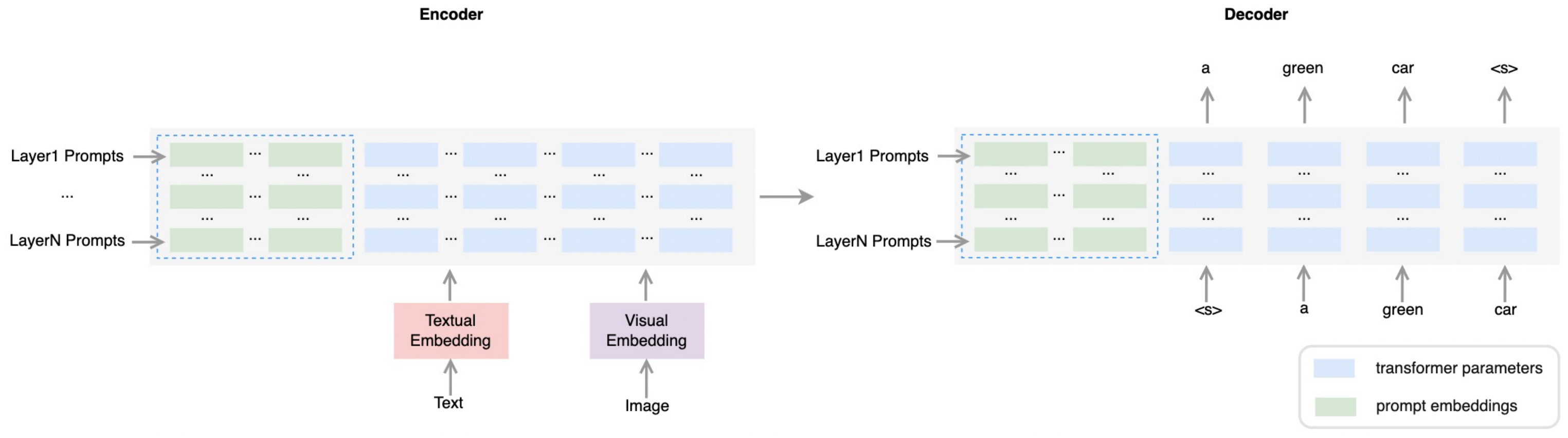}
    \caption{\textbf{Model overview. }An illustration of our multimodal prompt tuning architecture. Specifically, for the encoder and decoder, we add tunable prompt embeddings to each layer. }
    \label{fig:method}
\end{figure*}

This section introduces the details of our proposed method. It provides the detailed implementation of prompt tuning on a unified generative multimodal pretrained model. The overall framework is illustrated in Figure~\ref{fig:method}. 

\subsection{Preliminaries}

We select the unified sequence-to-sequence framework as it unifies understanding and generation tasks, and we specifically implement prompt tuning on the recent open-sourced state-of-the-art model OFA~\citep{ofa}.
In brief, it is built with a Transformer-based~\citep{transformer} encoder-decoder framework. 

Both the encoder and decoder consist of Transformer layers. To be more specific, an encoder layer consists of a multi-head self attention and a point-wise Feed-Forward Network (FFN). 
To build a connection between the encoder and decoder, the Transformer decoder layer additionally contains a cross-attention module in comparison with the encoder layer. 
The cross-attention is essentially multi-head attention, where the keys $K$ and values $V$ are the transformation of the encoder output states, instead of the inputs. 
Such architecture can handle tasks that provide inputs of the sequence-to-sequence format. 



In this work, we focus on prompt tuning for the transfer of the multimodal pretrained model. We leave the prompt learning in the stage of pretraining to the future work. 

\subsection{Prompt Tuning for Generative Multimodal Pretrained Models}
In the following, we introduce our implementation details of prompt tuning on the sequence-to-sequence multimodal pretrained model. Note that our method can extend to other generative multimodal pretrained models, e.g.,  BERT-like models. 

\paragraph{Basic Implementation} We focus on implementing prefix tuning~\citep{prefix_tuning, p_tuning_v2} based on its outstanding performance in either natural language understanding or generation. 
In comparison with the other prompt tuning methods, e.g., P-Tuning~\citep{p_tuning}, Prompt Tuning~\citep{prompt_tuning}, PPT~\citep{ppt}, adding soft prompt embeddings to each layer demonstrates enhanced training stability and improved downstream task performance even on relatively small models. 
Specifically, for the encoder and decoder, we add tunable prompt embeddings to each layer. 
Formally, we refer the pretrained model to a function $\mathcal{M}(\cdot)$, and the generation function of the prompt embeddings to $\mathcal{G}(\cdot)$. 
The formulation is demonstrated below:
\begin{align}
    y &= \mathcal{M}(\mathcal{G}(L, l), x),
\end{align}
where $x$ refers to the multimodal inputs, $L$ refers to the number of layers, and $l$ refers to the prompt length, which should be predefined by a hyperparameter. 
At each layer, we prefix soft prompt embeddings $p^{(i)}$ to the input hidden states $h^{(i)}$
Note that we only prefix prompt embeddings at Transformer layers. 
In the simplest practice, the prompt generator $\mathcal{G}$ is a sparse embedding matrix of $\mathbb{R}^{L \times l \times h}$, and we select the corresponding embedding at the $i$-th index and the $j$-th layer as the prompt embedding. Below we provide an illustration of some more complex implementations, and we compare those methods in this study. 

In the downstream tuning process, we only tune the newly added prompt embeddings at each layer and keep the parameters of the large pretrained model frozen. Therefore, while there are only a small amount of parameters that need to be updated, e.g., 1\%, the computation costs are far fewer than those of finetuning. 

\paragraph{Reparameterization} 
Except for the simplest implementation of adding a sparse embedding matrix at each layer, a more complex one should be adding an encoder, e.g., an MLP layer, to reparameterize prompt embeddings. 
We also investigate the influence of reparameterization in this context.

\paragraph{Prompt Length}
Similar to previous studies~\citep{prefix_tuning, p_tuning_v2}, we find that the length of prompt embeddings make a great difference in different downstream tasks. 
In this study, we investigate how this factor imposes influence on model performance in different downstream tasks. 

\paragraph{Prompt Depth}
To investigate the impacts of the place of prompt embedding insertion, we delve into the issue of prompt depth. Specifically, we simplify it to adding prompt embeddings to the encoder or decoder only, as well as to both modules. 

\section{Experiments}
\begin{table*}[t]
\center
\small
\vskip 0.15in
\begin{adjustbox}{max width=1.\textwidth}
\begin{tabular}{@{\extracolsep{\fill}}lcccccccccccccccc}
\toprule
  \multirow{2}*{Model}
  &\multicolumn{3}{c}{RefCOCO}
  &\multicolumn{3}{c}{RefCOCO+}
  &\multicolumn{2}{c}{RefCOCOg}
  &\multicolumn{2}{c}{SNLI-VE}
  &\multicolumn{4}{c}{COCO Captions}
  &\multicolumn{2}{c}{VQA}
 
  \\
  & val & testA & testB
  & val & testA & testB
  & val-u & test-u
  & dev & test
  & B@4 & M & C & S
  & test-dev & test-std
  \\
\midrule
    \multicolumn{9}{l}{\textit{Base}-size Models} \\
    Finetuning
    & 88.48 & 90.67 & 83.30
    & 81.39 & 87.15 & 74.29
    & 82.29 & 82.31
    & 89.30 & 89.20
    & 41.00 & 30.90 & 138.2 & 24.20
    & 78.00  & 78.10
    \\
    Prompt Tuning
    & 84.53 & 85.21 & 77.36
    & 76.34 & 81.44 & 67.68
    & 75.61 & 76.57
    & 88.18 & 88.59
    & 39.70 & 30.10 & 134.2 & 23.50
    & 74.31 & 74.47
    
    \\
\midrule
    \multicolumn{9}{l}{\textit{Large}-size Models} \\
    Finetuning
    & 90.05 & 92.93 & 85.26
    & 85.80  & 89.87 & 79.22
    & 85.89 & 86.55
    & 90.30  & 90.20
    & 42.40 & 31.50 & 142.2 & 24.50
    & 80.40  & 80.70
    \\
    Prompt Tuning
    & 90.05 & 92.31 & 85.59
    & 84.54 & 89.40 & 77.77
    & 85.27 & 85.89
    & 90.04 & 90.12
    & 41.81 & 31.51 & 141.4 & 24.42
    & 78.30  & 78.53
    \\
    
  
\bottomrule
\end{tabular}
\end{adjustbox}
\caption{Experimental results on RefCOCO, RefCOCO+, RefCOCOg, SNLI-VE, COCO Image Captioning, and VQA. For the base-size model, prompt tuning significantly underperforms finetuning, but for the large-size model, prompt tuning is able to achieve comparable performance. }
\label{tb:multimodal_results}
\end{table*}

To validate the effectiveness of prompt tuning for multimodal pretrained models, we conduct experiments on the conventional cross-modal tasks. 
Specifically, we experiment on cross-modal understanding and generation, including referring expression comprehension, visual entailment, image captioning, and visual question answering (VQA). 
We use the mostly-used base-size and large-size models for the experiments, whose sizes are around $180$M and $470$M respectively. We provide more details about the experimental setups in the Appendix~\ref{sec:appendix_setups}.

\subsection{Datasets \& Metrics}
\paragraph{Referring Expression Comprehension} 
We conduct experiments on the $3$ subtasks of referring expression comprehension, namely RefCOCO, RefCOCO+, and RefCOCOg~\citep{refcoco, refcocog}. 
This task requires the model to generate a correct bounding box that answers the given text query on a provided image. 
We use Acc@0.5 as the evaluation metric. 

\paragraph{Image Captioning}
We evaluate the image captioning capability of our method on the Microsoft COCO Image Captioning dataset~\citep{coco_cap}. 
In this task, the model should generate a description that corresponds to the information of the given image. 
We use BLEU@4~\citep{bleu}, METEOR~\citep{meteor}, CIDEr~\citep{cider}, and SPICE~\citep{spice} as the evaluation metrics. 

\paragraph{Visual Entailment}
To evaluate the performance of entailment, we implement the experiments on SNLI-VE~\citep{snli-ve}. 
Given an image and a text, the model should figure out their relations, whether they are entailment, contradiction, or neutrality. 
We follow the setups in \citep{ofa} and add the given premise to the input. We use accuracy as the evaluation metric.

\paragraph{VQA}
We implement our experiments on VQA 2.0~\citep{vqa, vqav2}. 
This task requires the model to generate the correct answer based on an image and a question about certain information on the image. 
Following \citet{ofa}, we use the all-candidate evaluation, which requires the model to generate a probability for each candidate among the $3,129$ most frequent answers. 
We use accuracy as the evaluation metric.

\subsection{Experimental Results}
Below we provide the detailed experiment results, including the comparison of prompt tuning and finetuning, as well as prompt tuning and other parameter-efficient tuning methods.

\begin{table*}[t]
\center
\small
\vskip 0.15in
\begin{adjustbox}{max width=1.0\textwidth}
\begin{tabular}{@{\extracolsep{\fill}}lcccccccccccccccc}
\toprule
  \multirow{2}*{Method}
  &\multicolumn{3}{c}{RefCOCO}
  &\multicolumn{3}{c}{RefCOCO+}
  &\multicolumn{2}{c}{RefCOCOg}
  &\multicolumn{2}{c}{SNLI-VE}
  &\multicolumn{4}{c}{COCO Captions}
  &\multicolumn{2}{c}{VQA}
 
  \\
  & val & testA & testB
  & val & testA & testB
  & val-u & test-u
  & dev & test
  & B@4 & M & C & S
  & test-dev & test-std
  \\

\midrule
    Bitfit
    & 89.61 & 92.20 & 84.91
    & 82.60 & 88.08 & 75.16
    & 84.66 & 84.68
    & 89.70 & 89.42
    & 41.02 & 30.92 & 138.8 & 24.23
    & 78.23  & 78.44
    \\
    
    Adapter
    & 90.01 & 92.30 & 85.02
    & 83.79 & 88.93 & 76.09
    & 85.10 & 85.45
    & 89.84 & 89.78
    & 41.38 & 31.16 & 139.5 & 24.30
    & 78.27  & 78.47
    \\
    
    Prompt Tuning
    & \textbf{90.05} & \textbf{92.31} & \textbf{85.59}
    & \textbf{84.54} & \textbf{89.40} & \textbf{77.77}
    & \textbf{85.27} & \textbf{85.89}
    & \textbf{90.04} & \textbf{90.12}
    & \textbf{41.81} & \textbf{31.51} & \textbf{141.4} & \textbf{24.42}
    & \textbf{78.30}  & \textbf{78.53}
    \\
\bottomrule
\end{tabular}
\end{adjustbox}
\caption{Evaluation of different parameter-efficient tuning methods using large-size models.  We find that prompt tuning can generally outperform Bitfit and Adapter. }
\label{tb:efficient tuning}
\end{table*}

\begin{figure*}[t]
    \centering
    \includegraphics[width=1.0\linewidth]{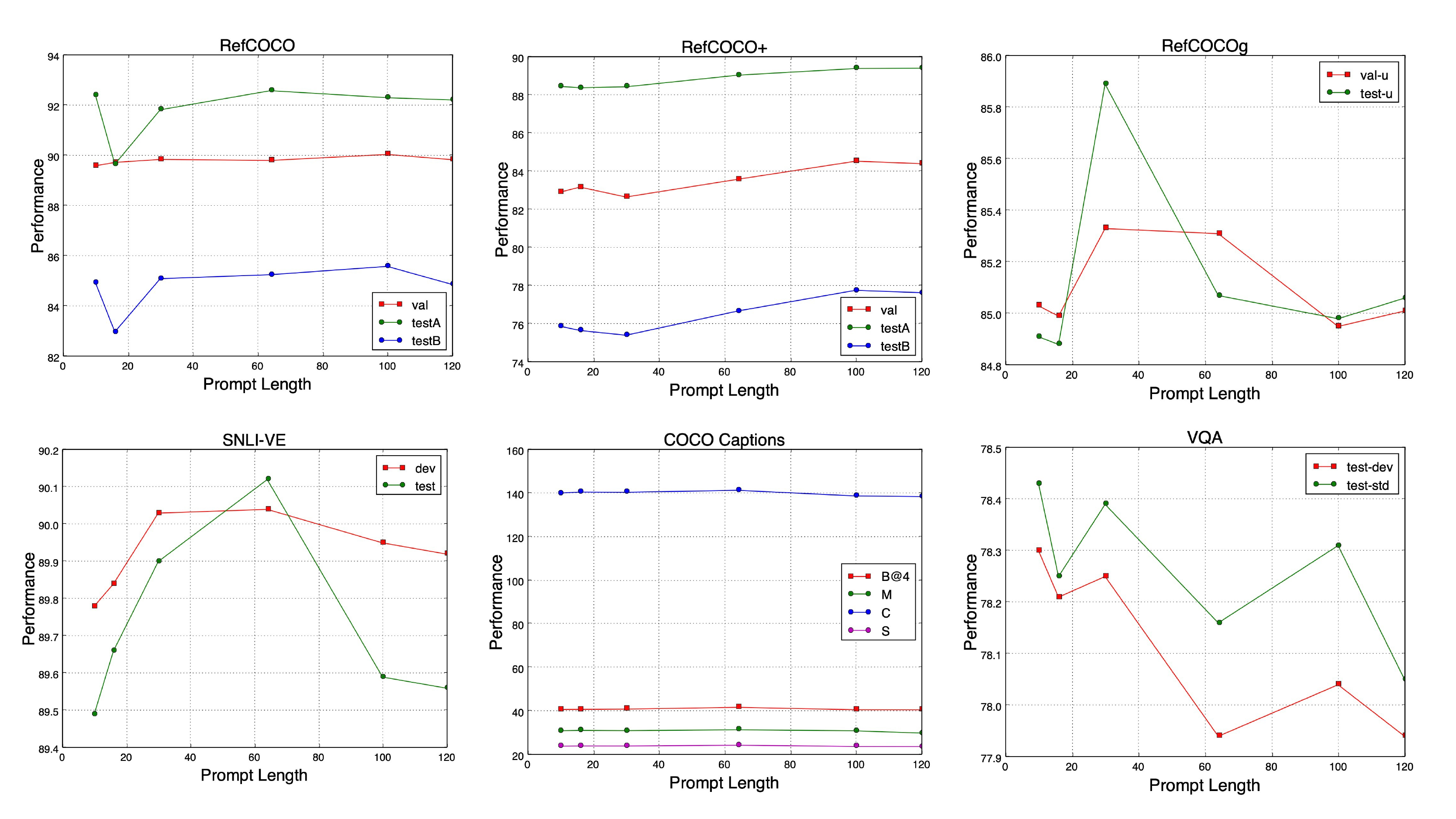}
    \caption{\textbf{Analysis of prompt lengths on multimodal downstream tasks. } We observe that increasing prompt lengths can generally bring performance improvements. Yet it cannot extend to all scenarios, and the increase might meet saturation. Based on the experimental results, we recommend $64$ for the prompt length as it helps the model achieve the average best results across tasks. }
    \label{fig:prompt_length}
\end{figure*}

\paragraph{Comparison with Finetuning} 
We demonstrate the experimental results of the $4$ tasks in Table~\ref{tb:multimodal_results}. 
In general, for the base-size model, prompt tuning underperforms finetuning by significant margins, but for the large-size model, prompt tuning is able to achieve comparable performance. 
To be more specific, in the evaluation of referring expression comprehension, for the base-size model, prompt tuning significantly underperforms finetuning by lagging behind a large margin of $5.64$ on average across RefCOCO, RefCOCO+, and RefCOCOg, but for the large-size model, prompt tuning only slightly underperforms finetuning by a small margin of $0.59$. 
In the evaluation of visual entailment, the gap between the algorithms is closer, which is around $0.17$. 
In the evaluation with the CIDEr score on image captioning, for the base-size model, prompt tuning underperforms finetuning by a margin of $4.0$, but for the large-size model, the performance gap is only $0.8$. 
In the evaluation of VQA, for the base-size model the performance gap is $3.63$ between prompt tuning and finetuning, and for the large-size model the gap is $2.17$ on the test-std set. Different from the other tasks, even in the experiments on the large-size model, the gap is still significant. 
We hypothesize that it is still necessary to search a better hyperparameter setup for this task due to the sensitivity of prompt tuning to hyperparameters.

\paragraph{Comparison with Other Parameter-Efficient Tuning Methods} 
We additionally add a comparison with two parameter-efficient tuning methods, namely Adapter~\citep{adapter} and BitFit~\citep{bitfit} to test whether prompt tuning is the best solution of light-weight transfer. 
Table~\ref{tb:efficient tuning} shows the results of different light-weight tuning methods implemented on the aforementioned datasets. 
In all the downstream tasks, prompt tuning surpasses the performance of Adapter and BitFit. 

\begin{figure*}[t]
    \centering
    \includegraphics[width=1.0\linewidth]{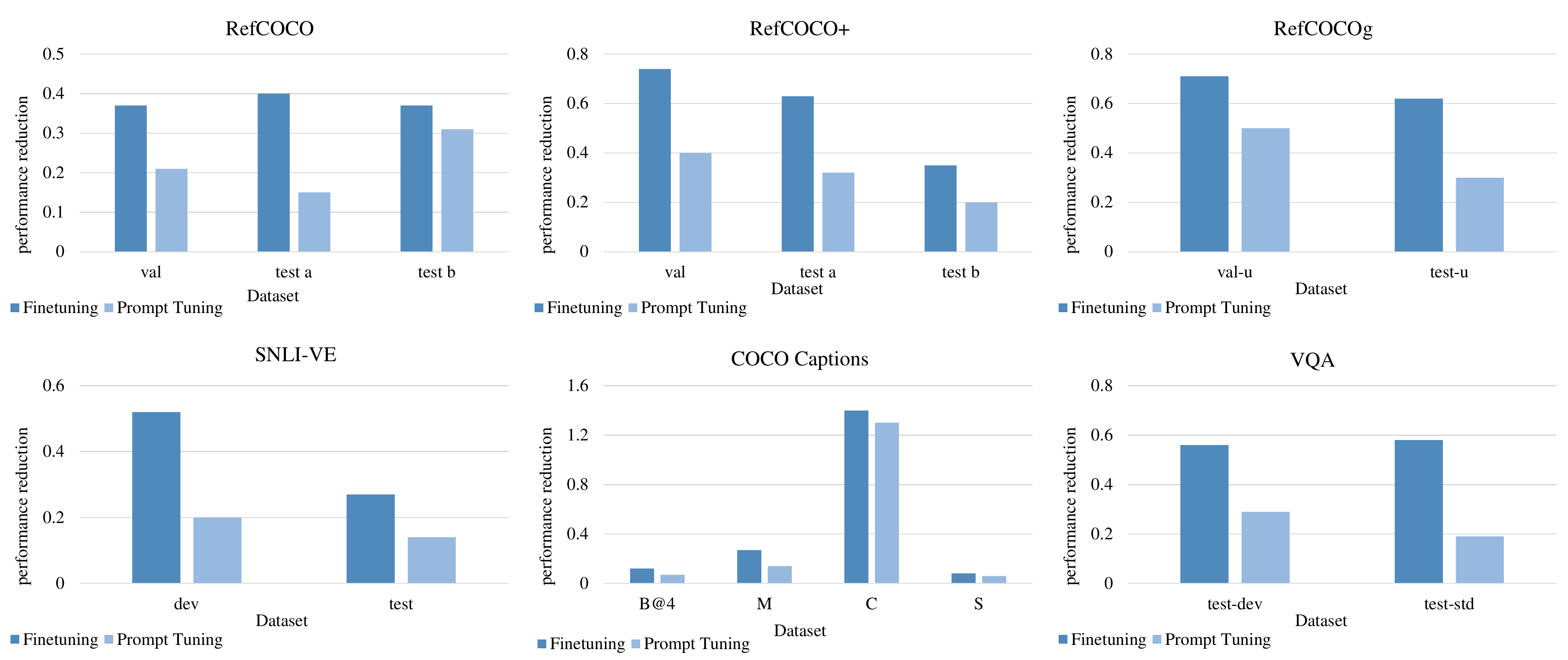}
    \caption{Experimental results on adversarial attack using large-size models. We discover that in the scenario of adversarial attack prompt tuning suffers from lower performance degradation across the tasks.}
    \label{fig:robustness}
\end{figure*}

\begin{table*}[t]
\center
\small
\vskip 0.15in
\begin{adjustbox}{max width=1.0\textwidth}
\begin{tabular}{@{\extracolsep{\fill}}lcccccccccccccccc}
\toprule
  \multirow{2}*{Method}
  &\multicolumn{3}{c}{RefCOCO}
  &\multicolumn{3}{c}{RefCOCO+}
  &\multicolumn{2}{c}{RefCOCOg}
  &\multicolumn{2}{c}{SNLI-VE}
  &\multicolumn{4}{c}{COCO Captions}
  &\multicolumn{2}{c}{VQA}
 
  \\
  & val & testA & testB
  & val & testA & testB
  & val-u & test-u
  & dev & test
  & B@4 & M & C & S
  & test-dev & test-std
  \\
\midrule

    Encoder
    & 89.48 & 91.71 & 84.98
    & 84.50 & 89.22 & 77.71
    & 85.07 & 85.58
    & 89.64 & 89.70
    & 41.39 & 31.08 & 141.1 & 24.34
    & 78.10  & 78.26
    \\
    
    Decoder
    & 88.90 & 91.28 & 84.32
    & 83.46 & 88.24 & 76.82
    & 84.54 & 85.02
    & 88.56 & 88.71
    & 40.08 & 30.43 & 140.8 & 24.06
    & 77.84  & 78.03
    \\
    
    Encoder+Decoder
    & 90.05 & 92.31 & 85.59
    & 84.54 & 89.40 & 77.77
    & 85.27 & 85.89
    & 90.04 & 90.12
    & 41.81 & 31.51 & 141.4 & 24.42
    & 78.30  & 78.53
    \\
\bottomrule
\end{tabular}
\end{adjustbox}
\caption{Evaluation of different prompt insertion methods. We specifically evaluate the performance of prompt tuning with prompts inserted to the encoder only, to the decoder only, or to both the encoder and decoder. }
\label{tb:prompt_depth}
\end{table*}

\begin{table*}[t]
\center
\small
\vskip 0.15in
\begin{adjustbox}{max width=1.0\textwidth}
\begin{tabular}{@{\extracolsep{\fill}}lcccccccccccccccc}
\toprule
  \multirow{2}*{Method}
  &\multicolumn{3}{c}{RefCOCO}
  &\multicolumn{3}{c}{RefCOCO+}
  &\multicolumn{2}{c}{RefCOCOg}
  &\multicolumn{2}{c}{SNLI-VE}
  &\multicolumn{4}{c}{COCO Captions}
  &\multicolumn{2}{c}{VQA}
 xqx
  \\
  & val & testA & testB
  & val & testA & testBxq
  & val-u & test-u
  & dev & test
  & B@4 & M & C & S
  & test-dev & test-std
  \\
\midrule

    Prompt Tuning
    & 90.05 & 92.31 & 85.59
    & 84.54 & 89.40 & 77.77
    & 85.27 & 85.89
    & 90.04 & 90.12
    & 41.81 & 31.51 & 141.4 & 24.42
    & 78.30  & 78.53
    \\
    
    MLP 
    & 90.12 & 92.56 & 85.63
    & 84.83 & 89.65 & 77.94
    & 85.42 & 86.01
    & 89.98 & 90.02
    & 41.67 & 31.48 & 140.7 & 24.40
    & 78.26  & 78.48
    \\
\bottomrule
\end{tabular}
\end{adjustbox}
\caption{Experimental results on reparameterization using large-size models. }
\label{tb:reparamerization}
\end{table*}

\subsection{Analyses}
In this section, we move forward to analyzing prompt tuning in multimodal pretraining. 
Specifically, we examine the robustness of prompt tuning based on the assumption that keeping most parameters of the pretrained model frozen should lead to improved robustness to adversarial attack. 
Also, we evaluate how different setups of prompt tuning, say the prompt length, the depth of prompt, and reparameterization, influence the downstream performance, and try to provide a recommended setup for consistently better performance. 

\paragraph{Robustness Analysis}
To test whether the pretrained model with prompt tuning for downstream transfer is robust, we conduct experiments of adversarial attack for the examination. 
Adversarial attack was first proposed in computer vision, which revealed the vulnerability of deep learning models. The most common adversarial attack methods in computer vision are gradient-based methods, such as FGSM~\citep{fgsm}, PGD~\citep{pgd}, MIM~\citep{mim} and SI~\citep{si}. 
Most of the typical unimodal adversarial attack on tasks are gradient-based methods. 
Among them, we select FGSM, which requires only one step of gradient computation on text and image embeddings. 
Experimental results are demonstrated in Figure~\ref{fig:robustness}. 
Prompt tuning consistently demonstrates better robustness in comparison with finetuning across all tasks. 
This confirms our hypothesis and also shows one significant advantage of prompt tuning not reflected in the standard evaluation. 
In practice, if model vulnerability is a issue that matters, we recommend the application of prompt tuning for the enhanced robustness without significant performance degradation. 

\paragraph{Prompt Length}
To study the effects of the prompt length on the final downstream performance, we evaluate the prompt tuning performance on the downstream tasks with a prompt length selected from $\{10, 16, 30, 64, 100, 120\}$. 
As shown in Figure~\ref{fig:prompt_length}, a general tendency is that a longer prompt length with more parameters to tune can encourage improvements in downstream performance across the tasks. 
However, we observe diminishing marginal utility and a prompt too long may even negatively impact the performance. 
Although the best prompt length for tasks are different, we empirically advise that the length of $64$ tokens can achieve a better performance on average. See Appendix~\ref{sec:appendix_results} for more details. 

\paragraph{Prompt Depth}
As we base our implementation on the encoder-decoder model, we intuitively assume that where to insert prompt embeddings matters the performance. 
To simplify this issue, in our practice, we evaluate the performance of inserting prompts to the encoder only, to the decoder only, or to both the encoder and decoder. 
Experimental results are demonstrated in Table~\ref{tb:prompt_depth}. 
We find that it is best to insert prompts to every layer of the whole Transformer model, though compared with the other alternatives it is less computation-efficient. 
In the comparison between insertion to the encoder only and to the decoder only, we observe that the former solution leads to a significantly better results across multiple downstream tasks. 
This suggests that the insertion of prompts to the bottom layers might contribute more to the success of downstream transfer. 

\paragraph{Reparameterization}
Empirically, directly updating the trainable embeddings leads to unstable optimization and a slight drop in performance. 
Prior work usually leveraged an encoder, e.g., an MLP~\citep{prefix_tuning}, to reparameterize the trainable embeddings. We evaluate the performance of reparameterization, and we demonstrate the experimental results in Table~\ref{tb:reparamerization}. For some datasets (e.g., RefCOCO and RefCOCOg), MLP brings consistent improvements.
For the others, MLP leads to relatively negative impacts (e.g., SNLI-VE and VQA). 
Thus we cannot come to a conclusion about which should be a preferable one. To achieve better performance on a specific dataset, it is still necessary to make an attempt on both methods.

\section{Discussion}
Apparently we have to admit that prompt tuning still cannot replace finetuning, and here we try to illustrate some of its limitations and point out some directions for future work. 

A salient problem is its slow convergence. 
Though prompt tuning has noticeable advantages in training efficiency, 
it costs at least $40$ epochs for prompt tuning to achieve the nearly best performance on some datasets (e.g., RefCOCO). 
A larger number of training epochs may incur more computation costs though prompt tuning has an advantage in training efficiency compared with finetuning. 
We demonstrate more details in Appendix~\ref{sec:appendix_results}. 
This indicates that finding a better solution for fast and stable convergence is also important besides reaching comparable or even improved performance over the conventional finetuning. 

Another common defect of prompt tuning is the difficulty in searching for a suitable hyperparamter setup. The hyperparameter tuning experience in finetuning is not suitable for prompt tuning. Fortunately, we find that prompt tuning for generative multimodal pretrained models is not as sensitive to hyperparameters as prompt tuning for pretrained language models. We provide details of hyperparameter setups in Appendix~\ref{sec:appendix_setups}. 

Despite the aforementioned limitations, prompt tuning demonstrates significantly better robustness against adversarial attack. 
In the future, we should pay more attention to this merit and find ways to leverage it.

\section{Conclusion}
In this work, we explore prompt tuning for generative multimodal pretrained models.  
Through extensive experiments, we demonstrate that the light-weight prompt tuning can achieve comparable performance with finetuning with much fewer parameters to tune (e.g., 1\%), and it can surpass other light-weight tuning methods, e.g., Adapter and BitFit. 
Through our analysis, we figure out a significant advantage of prompt tuning about its robustness against adversarial attack. 
Furthermore, we provide a comprehensive analysis about the influence of prompt tuning setups, including the prompt length, prompt depth, and reparameterization. 
Potentially prompt tuning can be an alternative to finetuning, but still, there are some salient limitations in this method, e.g., slow convergence and training instabilities. 
We hope that future studies in this field can alleviate the aforementioned problems and thus promote the application of prompt tuning.
\bibliography{anthology,custom}
\bibliographystyle{acl_natbib}

\appendix

\section{Appendix}
\label{sec:appendix}

\begin{figure}[t]
    \centering
    \includegraphics[width=.8\linewidth]{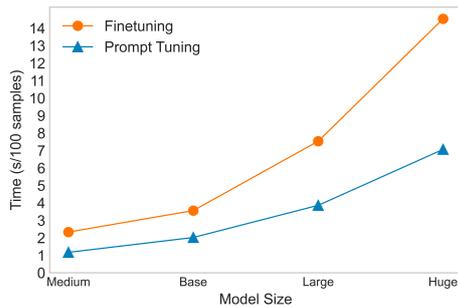}
    \caption{\textbf{Efficiency of different tuning methods. }We report the spent time per 100 samples of finetuning and prompt tuning on RefCOCO.}
    \label{fig:speed}
\end{figure}

\subsection{Experimental Setups}
\label{sec:appendix_setups}
\noindent \textbf{Referring Expression Comprehension}  Referring expression comprehension requires models to locate an image region described by a language query. We perform experiments on RefCOCO~\citep{refcoco}, RefCOCO+~\citep{refcoco}, and RefCOCOg~\citep{refcocog}. 
We report the standard metric Acc@0.5 on the validation and test sets. 
For finetuning, the batch
size is set to $128$, the learning rate is set to $0.03$, and the prompt length varies from $10$–$120$.

\noindent \textbf{Visual Entailment}  Visual entailment requires the model to evaluate the semantic relation between the given image and text, i.e., entailment, neutrality, or contradiction. We perform experiments on the SNLI-VE~\citep{snli-ve} dataset. 
We report accuracy on both dev and test sets. 
The model is finetuned  with a learning rate of $0.03$ and a batch size of $128$. The prompt length varies from $10$–$120$.

\noindent \textbf{Image Captioning}  Image captioning is a standard vision \& language task that requires models to generate an appropriate and fluent caption for an image. 
We report BLEU@4~\citep{bleu}, METEOR~\citep{meteor}, CIDEr~\citep{cider}, and SPICE~\citep{spice} scores on the Karpathy test split. 
We finetune the model with a learning rate of $0.03$, a batch size of $256$, and a prompt length varying from $10$–$120$. 
We only finetune the model with cross-entropy loss, without further CIDEr optimization. 

\noindent \textbf{Visual Question Answering}  Visual question answering~\citep{vqa, vqav2} is a cross-modal task that requires the models to answer the question given an image. 
We conduct experiments on VQA 2.0 and report the score on the test-std set. 
For finetuning, the batch size is set to $256$ and the learning rate is set to $0.03$. Exponential Moving Average (EMA) with a decay rate of $0.9999$ is employed in finetuning. The prompt length varies from $10$–$120$.

\subsection{Additional Experimental Results}
\label{sec:appendix_results}
In this section, we provide more experimental results for comprehensive understanding of the performance of prompt tuning. 

Below we summarize the detailed performance of prompt tuning on the downstream tasks in the conditions of different prompt lengths. See Table~\ref{tb:prompt_length}. On average, a prompt length of $64$ helps achieve the best average performance in the downstream tasks.

\begin{table}[t]
\center
\small
\vskip 0.15in
\begin{adjustbox}{max width=1.0\textwidth}
\begin{tabular}{@{\extracolsep{\fill}}lccccccc}
\toprule
 
  Method
  & Fintuning & Prompt Tuning
  
  \\
\midrule
    RefCOCO 
    & 40.00   
    & 77.44 
    \\
    
    SNLI-VE
    & 80.96  
    & 164.48   
    \\
    
    COCO Captions
    & 29.60   
    & 16.16   
    \\
    
    VQA
    & 616.16  
    & 455.52  
    \\
\bottomrule
\end{tabular}
\end{adjustbox}
\caption{Computation resource consumption of different tasks. We specifically compute the GPU-hours of both finetuning and prompt tuning on large-size models}
\label{tb:computation}
\end{table}

\begin{table}[t]
\center
\small
\vskip 0.15in
\begin{adjustbox}{max width=0.5\textwidth}
\begin{tabular}{@{\extracolsep{\fill}}lcccccc}
\toprule
 
  Length & 10 & 16 & 32 & 64 & 100 & 120
  \\
\midrule
  Score & 91.84 & 91.29 & 91.94 & 92.29 & 92.10 & 91.93
    
    
    
    \\
\bottomrule
\end{tabular}
\end{adjustbox}
\caption{Evaluation average performance of prompt tuning on the downstream tasks with different prompt lengths. }
\label{tb:prompt_length}
\end{table}

To evaluate the training efficiency of different methods, we experiment on the base model OFA of different sizes, spanning from $93$M to $930$M paramters. 
Figure~\ref{fig:speed} demonstrates their performance in efficiency by evaluating their used time of processing $100$ samples. 
We find that prompt tuning consistently performs better than finetuning in training efficiency. 
For the huge-size model, it can perform around $2$ times faster than finetuning. 
However, based on our observation, the advantage in training efficiency does not lead to less required computation resource. 
Table~\ref{tb:computation} lists the detailed computation resource consumption of both finetuning and prompt tuning.
Specifically, we compute the computation resource consumption by calculating the GPU-hours of finetuning and prompt tuning on different tasks. 
We find that for image captioning and VQA, prompt tuning consumes less resource, but for the other tasks prompt tuning adversely consumes more. 
It reflects that for tasks similar to pretraining tasks, especially those with more data in the pretraining stage, prompt tuning is able to outperform finetuning, but for others, prompt tuning even incurs more carbon footprints. 
This indicates that the real computation resource consumption for downstream transfer should be an important issue in the field of prompt tuning and the solution to this problem can further the developments of the application.


\end{document}